\documentclass[10pt, a4paper]{article}
\usepackage{lrec2022} 
\usepackage{multibib}
\newcites{languageresource}{Language Resources}
\usepackage{graphicx}
\usepackage{tabularx}
\usepackage{soul}
\usepackage{titlesec}
\titleformat{\section}{\normalfont\large\bfseries\center}{\thesection.}{1em}{}
\titleformat{\subsection}{\normalfont\SmallTitleFont\bfseries\raggedright}{\thesubsection.}{1em}{}
\titleformat{\subsubsection}{\normalfont\normalsize\bfseries\raggedright}{\thesubsubsection.}{1em}{}
\renewcommand\thesection{\arabic{section}}
\renewcommand\thesubsection{\thesection.\arabic{subsection}}
\renewcommand\thesubsubsection{\thesubsection.\arabic{subsubsection}}

\usepackage{epstopdf}
\usepackage[utf8]{inputenc}

\usepackage{hyperref}
\usepackage{xstring}

\usepackage{color}

\usepackage{times}
\usepackage{url}
\usepackage{latexsym}
\usepackage{booktabs}
\usepackage{multirow}
\usepackage{graphicx}
\usepackage{diagbox}
\usepackage{float}
\usepackage{amsmath}
\usepackage{amssymb}
\newcommand{\tabincell}[2]{\begin{tabular}{@{}#1@{}}#2\end{tabular}}

\usepackage{hyperref}
\usepackage{url}

\title{N24News: A New Dataset for Multimodal News Classification}

\name{Zhen Wang$^{1*}$, Xu Shan$^{2*}$, Xiangxie Zhang$^3$, Jie Yang$^4$}

\address{Delft University of Technology, Netherlands \\
         \{$^1$z.wang-42, $^3$x.zhang-60\}@student.tudelft.nl, \{$^2$x.shan-2, $^4$j.yang-3\}@tudelft.nl}

\abstract{
Current news datasets merely focus on text features on the news and rarely leverage the feature of images, excluding numerous essential features for news classification. In this paper, we propose a new dataset, \textit{N24News}, which is generated from New York Times with 24 categories and contains both text and image information in each news. We use a multitask multimodal method and the experimental results show multimodal news classification performs better than text-only news classification. Depending on the length of the text, the classification accuracy can be increased by up to 8.11\%. Our research reveals the relationship between the performance of a multimodal classifier and its sub-classifiers, and also the possible improvements when applying multimodal in news classification. \textit{N24News} is shown to have great potential to prompt the multimodal news studies. 
 \\ \newline  }

\begin{document}

\maketitleabstract

\def\thefootnote{*}\footnotetext{Equal Contribution}\def\thefootnote{\arabic{footnote}}

\section{Introduction}

People have tried to use different carriers to record news. Ancient people first drew images by hands-on walls to record important things. After language was invented, words became the main tools for recording. Thanks to parchment preserved to this day, we can study people who lived a long time ago. Later, with the invention of the camera, images are widely used in news. Compared with text, images can bring us more intuitive information, even if we cannot understand the language used in the news. It is safe to say that images and text play an equally important role in news. 

News classification is one of the essential tasks in news research \cite{katari2020survey}.
We use the information provided by the news to group them into different categories.
There is already some research about news classification, for example,  news datasets, such as 20NEWS \citelanguageresource{20news} and AG News \citelanguageresource{agnews}.
However, they choose to ignore the images and merely pay attention to the text.
This is not in line with the actual situation, especially when almost all the news today has images.
In this work, we aim to use both images and text to achieve better news classification.


\begin{figure*}[]
    \centering
    \includegraphics[width=1.0\textwidth]{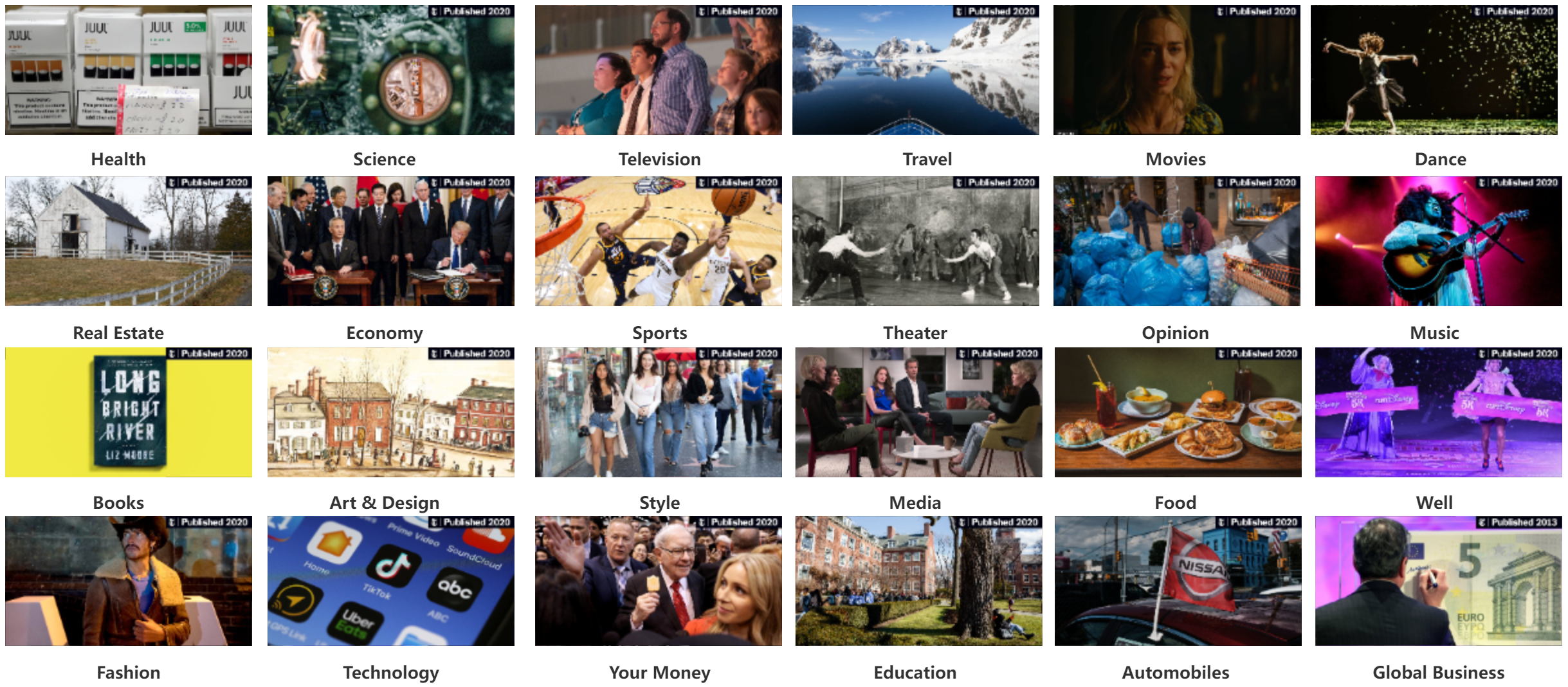}
    \caption{Image examples of 24 categories.}
    \label{fig:images}
\end{figure*}

%
%
In order to combine heterogeneous information extracted from images and texts, multimodal methods are needed. Multimodal approaches can process various types of information simultaneously and has been used in news studies before.
For example, in the fake news dataset Fakeddit \citelanguageresource{fakeddit}, the authors propose a hybrid text+image model to classifier fake news.
However, to best of our knowledge, currently there is no valid public news dataset containing enough real news with both images and texts that can be used to do multimodal news classification. 
Thus, in this work, we use the \textit{New York Times} to build a new dataset called \textit{N24News}. 
\textit{N24News} is a large-scale multimodal news dataset comprising 60K image-text pairs and 24 categories, which makes it possible to do multimodal real news classification tasks.
Further, we use a multitask multimodal network to conduct a preliminary experiment in multimodal news classification, and the experiment shows the multimodal method can achieve higher accuracy than text-only news classification.
Our error analysis reveals the relationship between the performance of a multimodal classifier and its sub-classifiers, and also the possible improvements when applying multimodal approaches in news classification.


\section{Related Work}

In news studies, the most commonly used datasets are 20NEWS \citelanguageresource{20news} and AG NEWS \citelanguageresource{agnews}. 20NEWS is a collection of approximately 20,000 newsgroup documents across 20 different newsgroups, and AG News contains 1 million news articles gathered from more than 2000 news sources and grouped into four categories. These two datasets are now used as benchmarks for testing text classification models, such as BERT \cite{devlin2018bert} and XLNet \cite{yang2019xlnet}. 

Multimodal deep learning \cite{ngiam2011multimodal} is able to leverage different types of features, such as voice, image, and text, to achieve better performance. 
Nowadays, multimodal methods have been used in lots of tasks, for example, multimodal sentiment analysis \cite{soleymani2017survey}, multimodal translation \citelanguageresource{sanabria2018how2}, multimodal emotion recognition \cite{tzirakis2017end}, and multimodal question answering \citelanguageresource{yagcioglu2018recipeqa}. 

One common multimodal architecture is to use different types of models to process the corresponding input data, such as first using an image classifier to obtain image features, a text classifier to obtain text features, and then combining these features before subsequent processing. In multimodal deep learning, the most critical part is feature fusion. 
Recent researches have proposed various feature fusion methods \cite{zhang2020multimodal}. \textit{Concatenation} \cite{nojavanasghari2016deep,anastasopoulos2019neural} is the most commonly used method. It splices different features directly along a certain dimension.
Further, other fusion methods, for example, \textit{weighted-sum} and \textit{pooling}, are also able to achieve good results.
\textit{Weighted-sum} \cite{vielzeuf2018centralnet} assigns different weights to each feature and sum them up. \textit{Pooling} \cite{chao2015long} methods, including \textit{max-pooling} and \textit{average-pooling}, are also used in many fusion scenarios, which can find the most important pieces of information in each feature and finally integrate them.
Additionally, \textit{attention-based} fusion methods \cite{zhang2020tell,shih2016look}, which using the attention mechanism to let the model learn to automatically find the most crucial part of the feature through training, are playing an increasingly important role in multimodal deep learning tasks.

\begin{table}[]
\centering
\begin{tabular}{lrlr}
\toprule
\textbf{Category}    & \multicolumn{1}{l}{\textbf{Count}} & \textbf{Category}         & \multicolumn{1}{l}{\textbf{Count}} \\\hline
Health      & 3000  & Books            & 3000  \\
Science     & 3000  & Art \& Design    & 3000  \\
Television  & 3000  & Style            & 2681  \\
Travel      & 3000  & Media            & 3000  \\
Movies      & 3000  & Food             & 3000  \\
Dance       & 3000  & Well             & 681   \\
Real Estate & 3000  & Fashion          & 3000  \\
Economy     & 1761  & Technology       & 3000  \\
Sports      & 3000  & Your Money       & 1263  \\
Theater     & 3000  & Education        & 825   \\
Opinion     & 3000  & Automobiles      & 1825  \\
Music       & 3000  & Global Business  & 1182  \\\bottomrule
\end{tabular}
\caption{Statistics of 24 categories.}
\label{tab:statistics}
\end{table}

%
Multimodal methods are also commonly used in news studies.
Previous multimodal news researches mainly focus on fake news detection. 
\newcite{fakeddit} propose a multimodal fake news dataset from Reddit with six categories according to the degree and type of fake news in the news. 
\newcite{giachanou2020multimodal} use \textit{word2vec} to extract the news text features and five different image models to extract news image features. \newcite{wang2018eann} use an adversarial neural network to identify fake news on newly emerged events in online social platforms. 
Fake news detection is a variant of news classification, which mostly has binary categories (true or false), making the task is not so difficult. Furthermore, there are few studies on the application of multimodal classification focus on real news.

In that case, we collect and apply multimodal methods on our dataset \textit{N24News}, which containing massive news images and texts, as well as many different categories, to facilitate the research of multimodal news classification applied in real news study. The code and dataset will be on Github \footnote{\url{https://github.com/billywzh717/N24News}}.

\begin{table}[]
\centering
\begin{tabular}{|p{7.2cm}|}
\hline
\textbf{Category}: Movies \\
\textbf{Headline}: A Man’s Death, a Career’s Birth \\
\textbf{Image}: \\
$\includegraphics[width=0.25\textwidth]{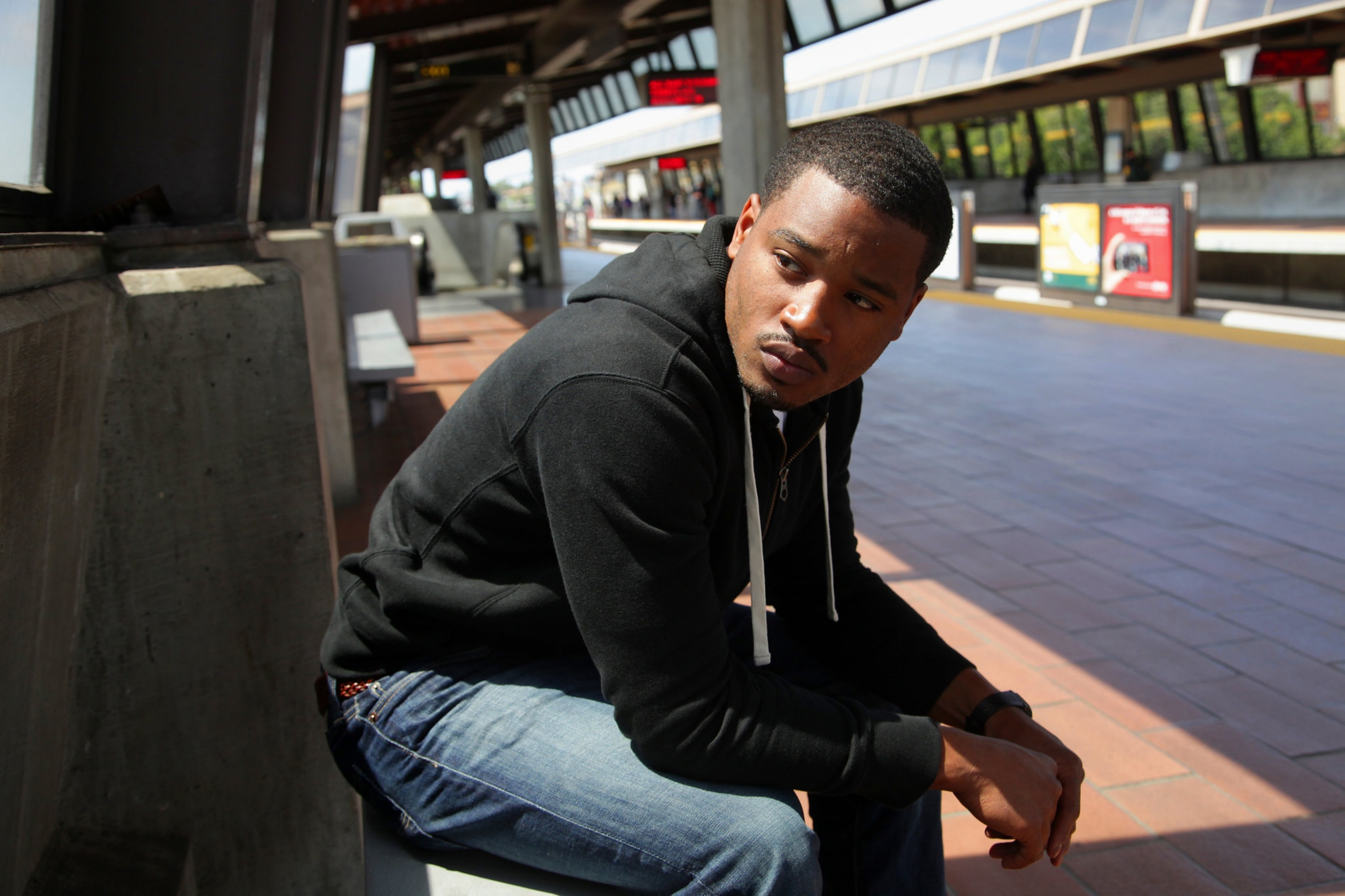}$ \\
\textbf{Caption}: Ryan Coogler on the BART platform at Fruitvale, where Oscar Grant III was killed. His film of that story, “Fruitvale Station,” opens next month. \\
\textbf{Abstract}:A killing at a Bay Area rapid-transit station has inspired Ryan Coogler’s feature-film debut, a movie already honored at the Sundance and Cannes film festivals. \\
\textbf{Body}: OAKLAND — It had been nearly a year since Ryan Coogler last stood on the arrival platform on the upper-level of the Fruitvale Bay Area Rapid Transit Station, where 22-year-old Oscar Grant III, unarmed and physically restrained, was shot in the back by a BART transit officer... \\ \hline
\end{tabular}
\caption{A sample from \textit{N24News}.}
\label{tab:example}
\end{table}

\section{The N24News Dataset}

\subsection{Dataset Collection}
The \textit{N24News} is extracted from the New York Times. New York Times is an American daily newspaper that was founded in 1851. It publishes worldwide news on various topics every day. Starting from the 2000s, the New York Times fully turned to digitization \cite{perez2008times}, and previous news was transferred to the Internet to facilitate people's reading and provide internet API for scientific research purposes.

To build the \textit{N24News} dataset, we use the API provided by New York Times to obtain all the links published from 2010 to 2020. Then we use these links to retrieve all the actual web pages in the past decade. After analyzing those web pages, we exclude video news, and only the news articles in text form are retained. While most news has only one image, to better balance the number of images and news, we only choose one image for each news and drop out the news which does not contain any images.
All news belongs to 24 different categories. We do not merge similar categories, such as science and technology, arts and theater. To make the dataset more balance, we collect up to 3000 samples for each category.
Finally, 60K news articles are collected in total. The amount of each category is shown in Table \ref{tab:statistics}. Each article sample contains one category tag, one headline, one abstract, one article body, one image, and one corresponding image caption. An example is shown in Table \ref{tab:example}. 
We randomly split datasets into training/validation/testing sets in the ratio of 8:1:1. Compared with other multimodal research such as fake news detection, our dataset comes from a professional news website, which ensures the correctness of the dataset, thus no additional manual annotation work is needed.


\subsection{Dataset Statistics}

\begin{table*}[]
\centering
\begin{tabular}{@{}llllll@{}}
\toprule
\textbf{Dataset}                     & \textbf{Size}      & \textbf{Classes} & \textbf{Type}        & \textbf{Source}                 & \textbf{Topic}  \\ \midrule
20NEWS \citelanguageresource{20news}                      & 20,000    & 20      & text        & Newsgroup              & real news \\
AG NEWS \citelanguageresource{agnews}                    & 1,000,000 & 4       & text        & AG News                & real news \\
Guardian News \citelanguageresource{guardiannews}              & 52,900    & 4       & text        & Guardian News & real news \\
Yahoo News \citelanguageresource{yahoonews}                 & 160,515    & 31      & text        & Yahoo                  & real news \\
BBC News \citelanguageresource{bbcnews} & 2,225     & 5       & text        & BBC                    & real news \\\hline
\begin{tabular}[c]{@{}l@{}} BreakingNews \\ \citelanguageresource{breakingnews} \end{tabular} & 110,000   & none      & text, image & RSS Feeds              & real news \\
\begin{tabular}[c]{@{}l@{}} TREC Washington Post \\ \citelanguageresource{washingtonpost} \end{tabular} & 728,626   & none      & text, image & Washington Post & real news \\
Fauxtography \citelanguageresource{fauxtography}               & 1,233     & 2       & text,image  & Snopes, Reuters        & fake news \\
\begin{tabular}[c]{@{}l@{}} Image-verification-corpus \\ \citelanguageresource{imageverificationcorpus} \end{tabular}     & 17,806    & 2       & text,image  & Twitter                & fake news \\
Fakeddit \citelanguageresource{fakeddit}               & 1,063,106 & 2,3,6   & text,image  & Reddit                 & fake news \\\hline
\textbf{N24News (Ours)} & \textbf{61,218}   & \textbf{24} & \textbf{text, image} & \textbf{New York Times} & \textbf{real news} \\ \bottomrule
\end{tabular}
\caption{Comparison of various news datasets.}
\label{tab:compare}
\end{table*}

In Table \ref{tab:compare}, we show some information about \textit{N24News} and other news-related datasets in previous researches. Compare to other datasets, \textit{N24News} has some unique advantages. 
Firstly, \textit{N24News} has 24 categories, which exceeds most of the previous news datasets, especially compare with multimodal datasets.
Moreover, there is no valid multimodal news dataset that can be used to do real news classification before \textit{N24News}. Previous multimodal researches in news classification mainly focus on fake news detection with limited categories.
The lengths of \textit{Headline}, \textit{Caption}, \textit{Abstract} and \textit{Body} are 52.33, 115.27, 129.42 and 4701.08 respectively. From \textit{Headline} to \textit{Body}, average lengths are progressively increasing.
This allows us to study the gain effect of images on different lengths and different types of text classification tasks with \textit{N24News}.

\begin{figure}[]
    \centering
    \includegraphics[width=0.49\textwidth]{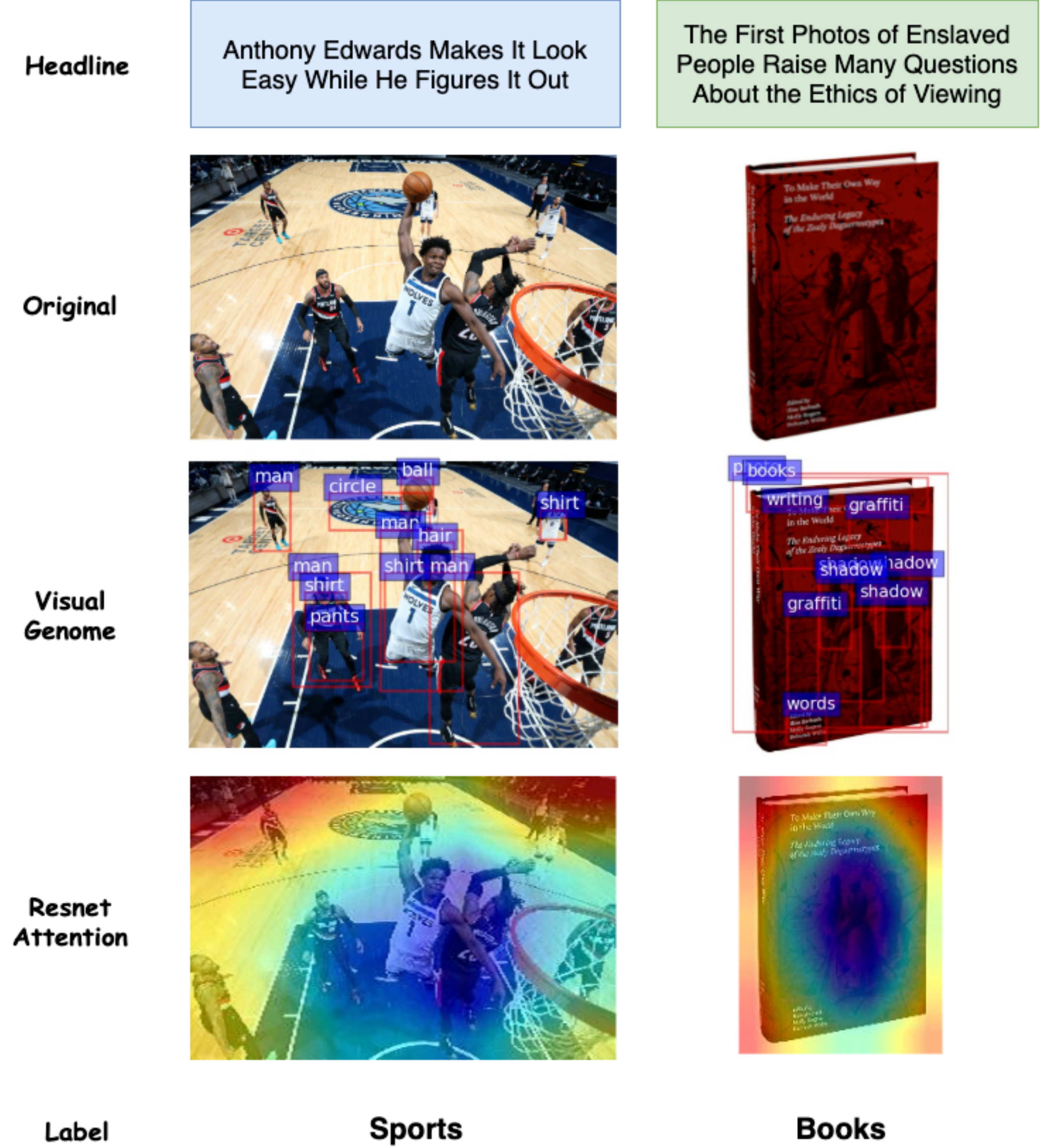}
    \caption{Visualization of critical parts in images. The news headlines lacks keywords that can be used for classification, but there are in the images.}
    \label{fig:vg}
\end{figure}

\subsection{Multimodal Analysis}

Text feature in classification task has been well studied before, so we will focus on what the news images in \textit{N24News} are able to provide to improve the classification results.
In Figure \ref{fig:images}, we list some image examples of each category. It is obvious that news images are usually closely related to the category they belong to. 

To better understand what can be learned from news images by current image classification models, we use a Faster-RCNN \cite{fasterrcnn} model trained on Visual Genome \cite{krishnavisualgenome}, a dataset aiming at providing semantic information from images.
We also use a Resnet\cite{resnet} trained on \textit{N24News} to reveal the critical part of news images. 
Two examples are shown in Figure \ref{fig:vg}. Faster-RCNN extracts the important semantic information in the images, such as ball and books, and Resnet focuses on salient objects: player and book cover. 
In Figure \ref{fig:vg}, it is hard to recognize the topic only given the two headlines.
The one on the left-hand side may be related to many topics, while the right-hand side one is closer to the topic of \textit{opinion}. However, with the information obtained by images, we can easily guess that the left one is about a sports-related topic, while the right one is about a book. Figure \ref{fig:vg} shows that the information in the image can be used to strengthen news classification.

Moreover, The information provided by images can also help to distinguish similar categories. 
There are limited similar categories in previous news datasets. However, in \textit{N24News}, there are some similar categories, for example, theater and movies. Only with text, it is difficult to tell the story is happening in a theater or on a screen, but images make things much easier. Theater-related images always happened on a stage, but movies not, as shown in Figure \ref{fig:images}. This makes a huge difference, and if we can make good use of image information, the classification accuracy will be much higher.

\subsection{Challenges}

\begin{figure}[]
    \centering
    \includegraphics[width=0.4\textwidth]{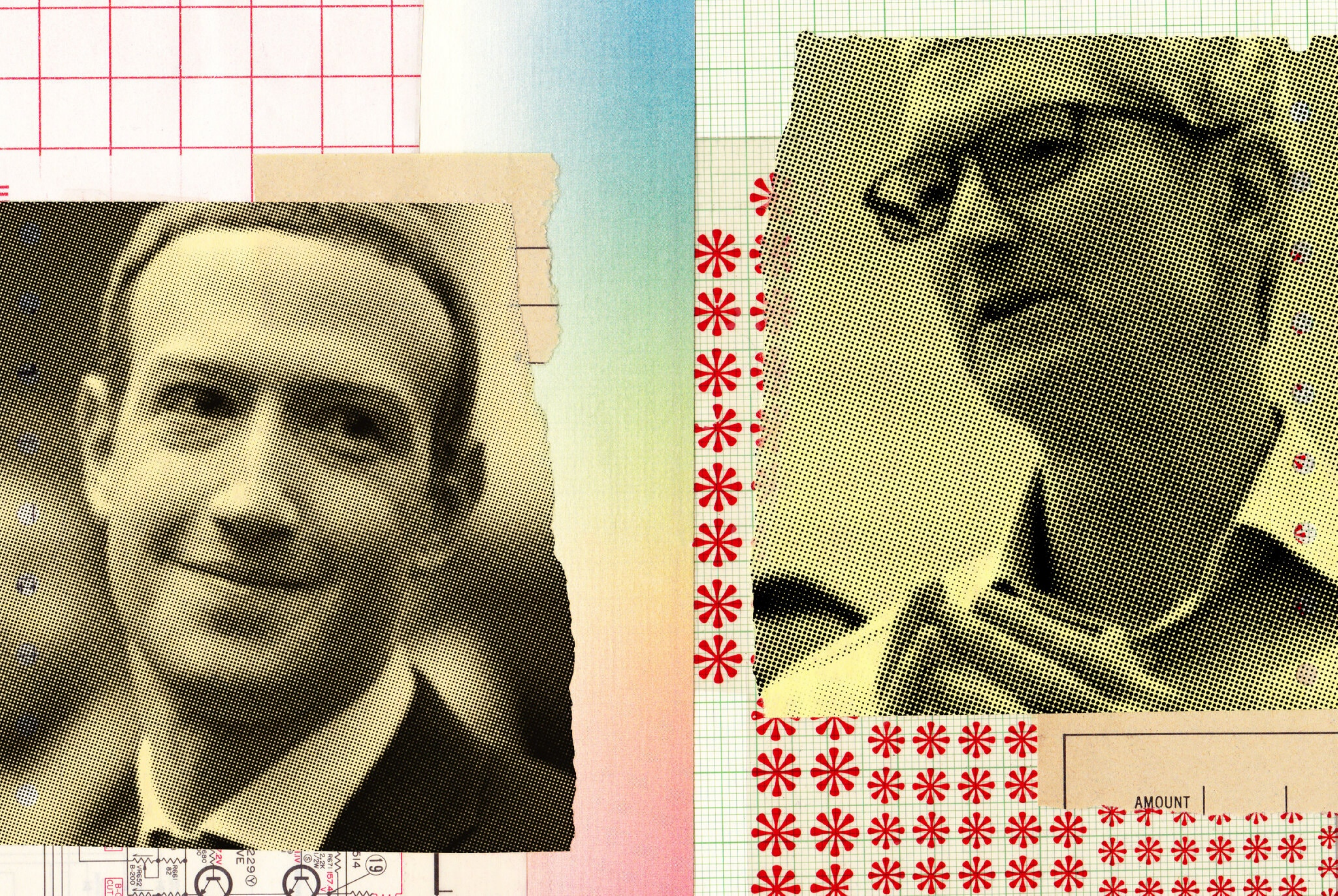}
    \caption{\textit{Breaking Point: How Mark Zuckerberg and Tim Cook Became Foes}}
    \label{fig:context}
\end{figure}

While multimodal data can introduce lots of new information to facilitate the news classification, \textit{N24News} also releases some new challenges. The biggest challenge is how to better understand news images. 
Current image classification models are able to extract the features of objects and the relationships between them in images. 
However, directly using those models to classify news images cannot achieve a strong result because they are mainly designed to classify specific objects, such as cats or dogs, while a news image is more likely to reflect an event. An identical object may have different meanings in different scenarios. For example, the two people in Figure \ref{fig:context} are Facebook and Apple’s CEO, thus this image comes from a news related to technology. However, existing image models only recognize there are two people but cannot obtain more meaningful information. Therefore, the features obtained through those models cannot fully reflect the hidden contextual information in news images. We hope \textit{N24News} can also prompt the research in event image classification, which is a new and challenging field.


\section{Model}

\begin{figure}[]
    \centering
    \includegraphics[width=0.48\textwidth]{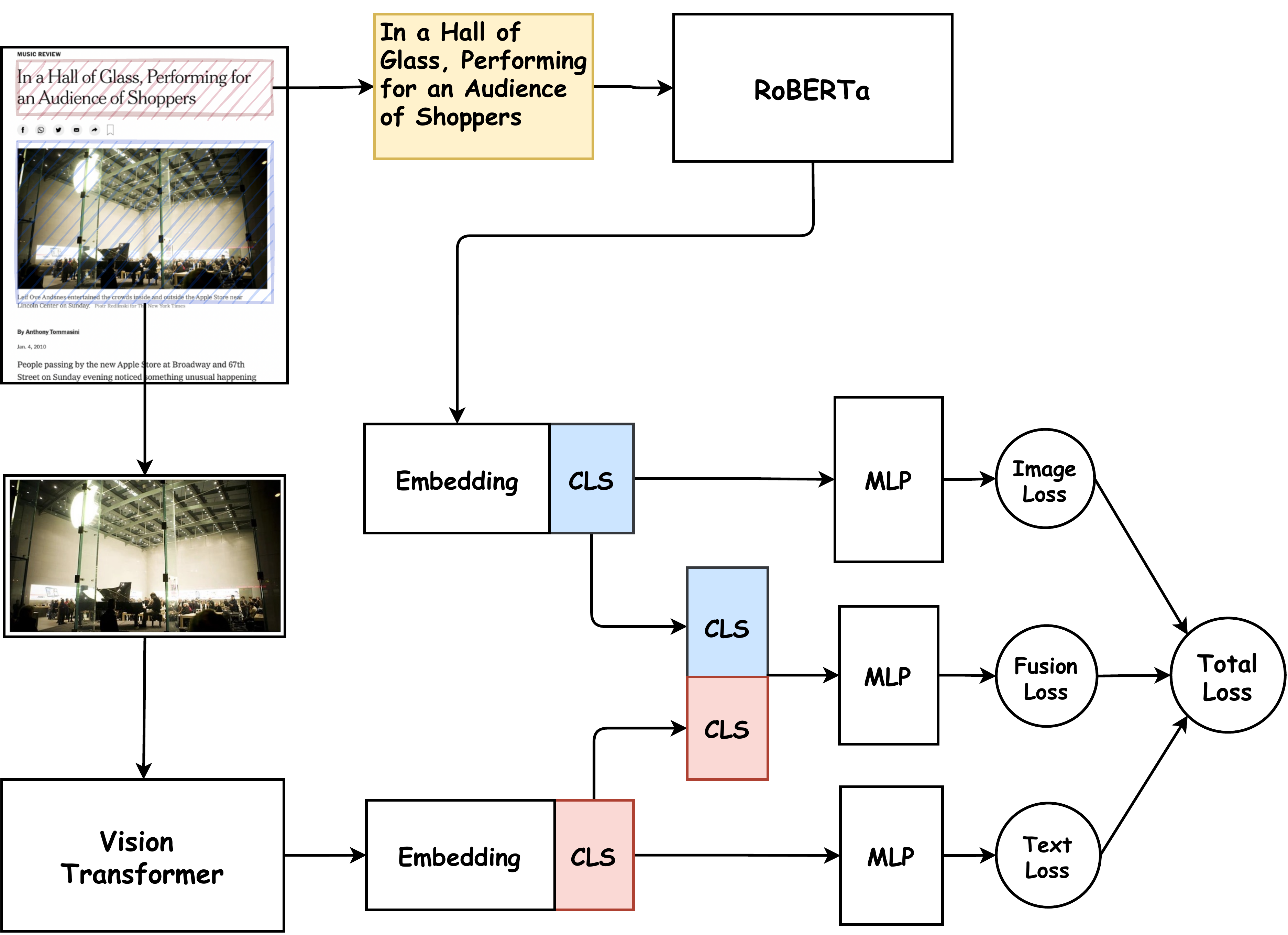}
    \caption{Overview of our multitask multimodal network.}
    \label{fig:mlp}
\end{figure}

To figure out how images can enhance the news classification and the potential challenges when applying the multimodal methods, we use a simple multitask multimodal network and conduct some experiments on \textit{N24News}.
As illustrated in Figure \ref{fig:mlp}, our model consists of two kinds of feature extraction models. On the bottom is Vision Transformer (ViT) \cite{dosovitskiy2020image}, one of the current state-of-the-art image classification models. Above it is RoBERTa\cite{liu2019roberta}, one of the current state-of-the-art text classification models. The ViT we use is pre-trained on imagenet2012 \cite{krizhevsky2012imagenet}, consists of a Resnet-50 and a base version of vision transformer with 12 layers transformer encoder. The pre-trained RoBERTa is also a base version and consists 12 layers transformer encoder.

We firstly use ViT and RoBERTa to obtain the image feature and text feature separately, where $embedding$ is the embeddings extracted from the original text and image, $CLS$ is a 1D embedding containing the information of its corresponding image or text. Then we concatenate those two kinds of features together. After obtaining the fused feature, we then use three multilayer perceptrons (MLPs) to predict the label for image feature, text feature, fusion feature separately. Finally, the cross-entropy is used to calculate the $Loss$ for each prediction. The final $Total Loss$ will be the sum of all three types of $Loss$. When testing on the test set, we only use the output of the fusion feature to calculate the final prediction result.

\begin{table}[]
\centering
\begin{tabular}{llll}
\toprule
\multicolumn{2}{c}{\textbf{Modal Type}} & \textbf{F1} & \textbf{ACC} \\\hline
\multicolumn{4}{c}{\textbf{Image Only}}                               \\\hline
Image                 & \multicolumn{1}{c}{-} &  52.80  & 54.34 \\\hline
\multicolumn{4}{c}{\textbf{Text Only}}                               \\\hline
\multicolumn{1}{c}{-} & Headline              &  70.31  & 71.98    \\
\multicolumn{1}{c}{-} & Caption               &  71.56  & 73.87    \\
\multicolumn{1}{c}{-} & Abstract              &  78.19  & 79.65    \\
\multicolumn{1}{c}{-} & Body               &  87.65  & 88.86    \\\hline
\multicolumn{4}{c}{\textbf{Multimodal}}                               \\\hline
Image                 & Headline              &  78.42  & 79.41    \\
Image                 & Caption               &  76.33  & 77.45    \\
Image                 & Abstract              &  82.52  & 83.33    \\
Image                 & Body               &  90.44  &  91.08   \\
\bottomrule
\end{tabular}
\caption{The evaluation results on the \textit{N24News} testing set.}
\label{tab:experiments}
\end{table}

\begin{table*}[]
\centering
\begin{tabular}{@{}lllllll@{}}
\toprule
\multicolumn{3}{c}{\textbf{Predict Result}} & \multirow{2}{*}{\textbf{Percent}} & \multicolumn{3}{c}{\textbf{Example}} \\ \cmidrule(r){1-3} \cmidrule(r){5-7}
\textbf{Multi} & \textbf{Text} & \textbf{Image} & & \multicolumn{1}{l}{\textbf{Image}} & \multicolumn{1}{l}{\textbf{Headline}} & \multicolumn{1}{l}{\textbf{Label}} \\ \midrule
\multirow{3}{*}{True}  & \multirow{3}{*}{True}  & \multirow{3}{*}{True} & \multirow{3}{*}{42.46\%}   & \multirow{3}{*}{$\includegraphics[width=0.12\textwidth]{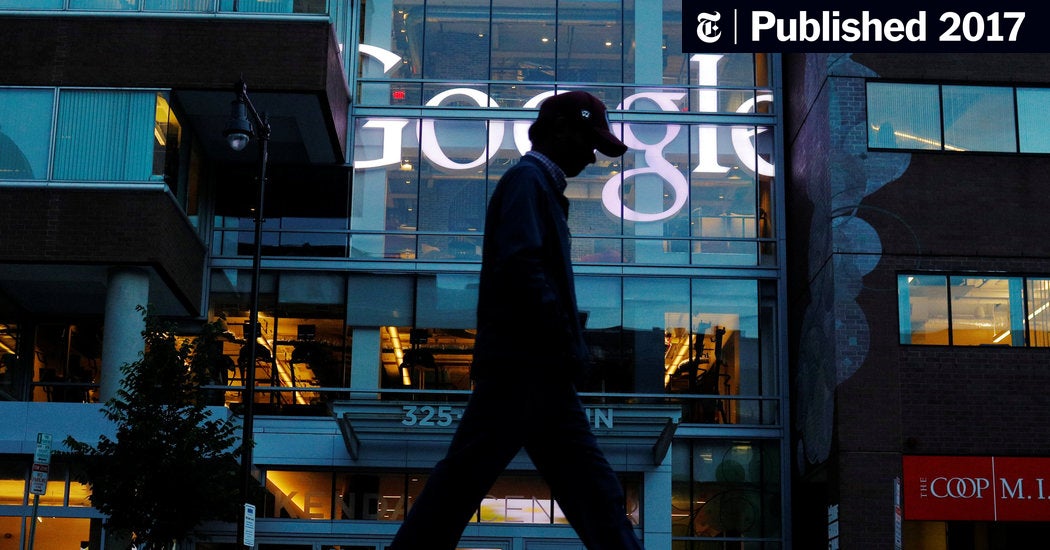}$} & \multirow{3}{*}{\tabincell{l}{For Alphabet, a Record \\ Fine Is Both a Footnote \\ and a Warning}} & \multirow{3}{*}{Technology} \\
& & & & & & \\
& & & & & & \\ \midrule
\multirow{3}{*}{True}  & \multirow{3}{*}{False} & \multirow{3}{*}{False}   & \multirow{3}{*}{2.56\%} & \multirow{3}{*}{$\includegraphics[width=0.12\textwidth]{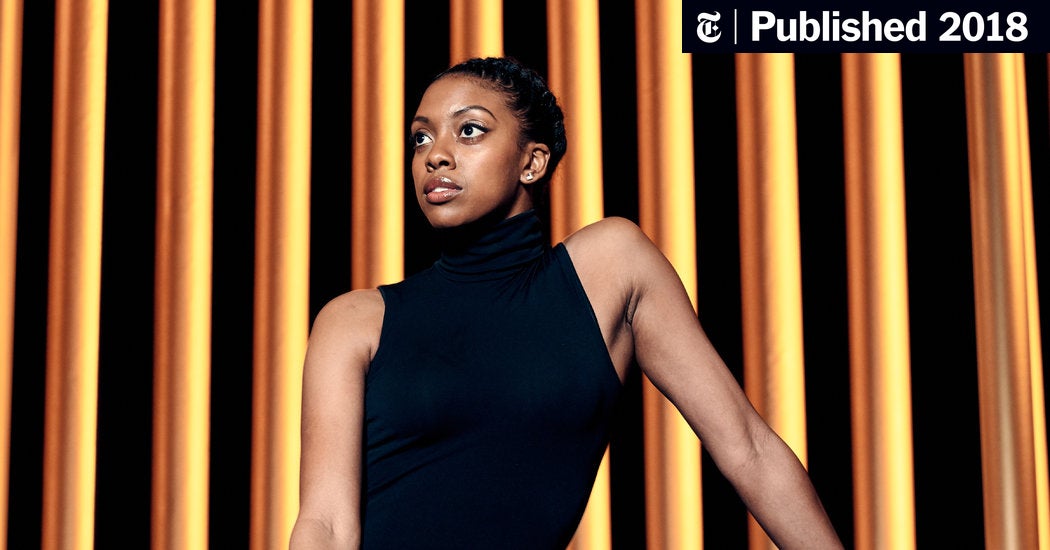}$}  & \multirow{3}{*}{\tabincell{l}{Joan of Arc, Superstar? \\ Not to the Woman Playing \\ Her}} & \multirow{3}{*}{Theater} \\
& & & & & & \\
& & & & & & \\\midrule
\multirow{3}{*}{True}  & \multirow{3}{*}{False}  & \multirow{3}{*}{True} & \multirow{3}{*}{7.01\%} & \multirow{3}{*}{$\includegraphics[width=0.12\textwidth]{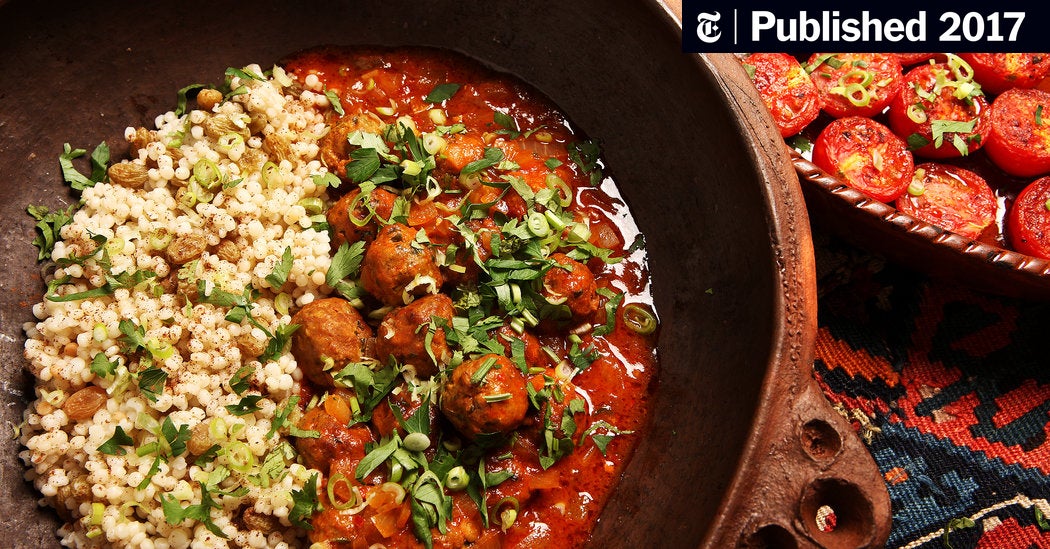}$} &  \multirow{3}{*}{\tabincell{l}{Time to Shift Gears}} & \multirow{3}{*}{Food} \\
& & & & & & \\
& & & & & & \\\midrule
\multirow{3}{*}{True}  & \multirow{3}{*}{True} & \multirow{3}{*}{False} & \multirow{3}{*}{27.69\%}   & \multirow{3}{*}{$\includegraphics[width=0.12\textwidth]{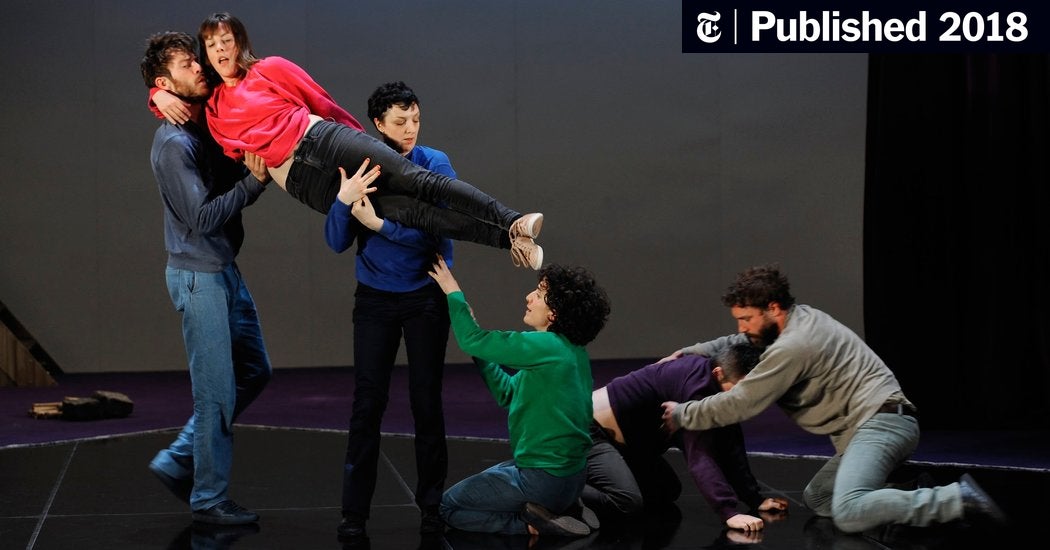}$} & \multirow{3}{*}{\tabincell{l}{10 Dance Performances to \\ See in NYC This Weekend}} & \multirow{3}{*}{Dance}    \\
& & & & & & \\
& & & & & & \\\midrule
\multirow{3}{*}{False} & \multirow{3}{*}{True}  & \multirow{3}{*}{True}  & \multirow{3}{*}{0.03\%}  & \multirow{3}{*}{$\includegraphics[width=0.12\textwidth]{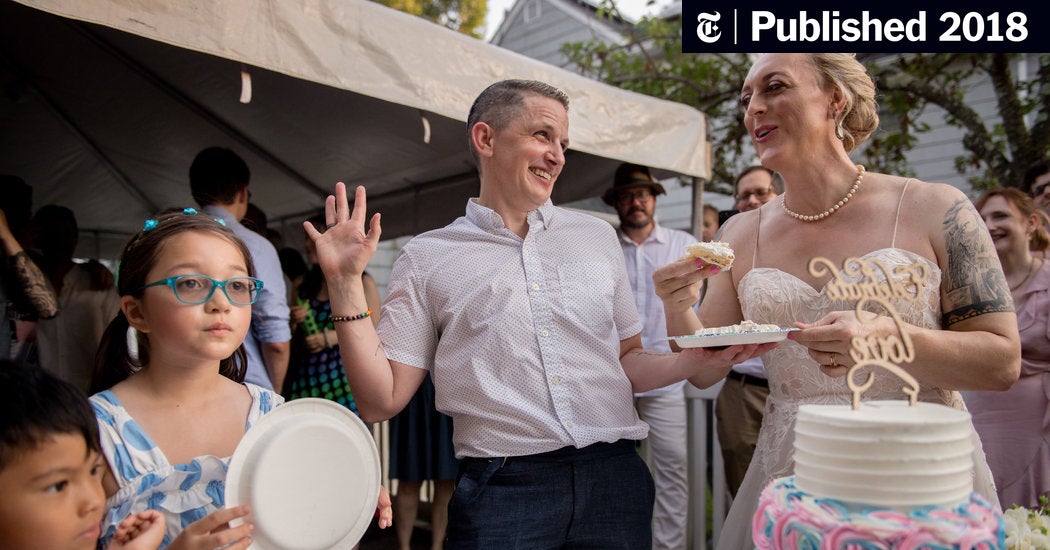}$}  & \multirow{3}{*}{\tabincell{l}{His Dating Profile Listed \\ Reasons Not to Date Him. \\ She Was Intrigued.}} & \multirow{3}{*}{Fashion}    \\
& & & & & & \\
& & & & & & \\\midrule
\multirow{3}{*}{False} & \multirow{3}{*}{False}  & \multirow{3}{*}{True} & \multirow{3}{*}{3.63\%} & \multirow{3}{*}{$\includegraphics[width=0.12\textwidth]{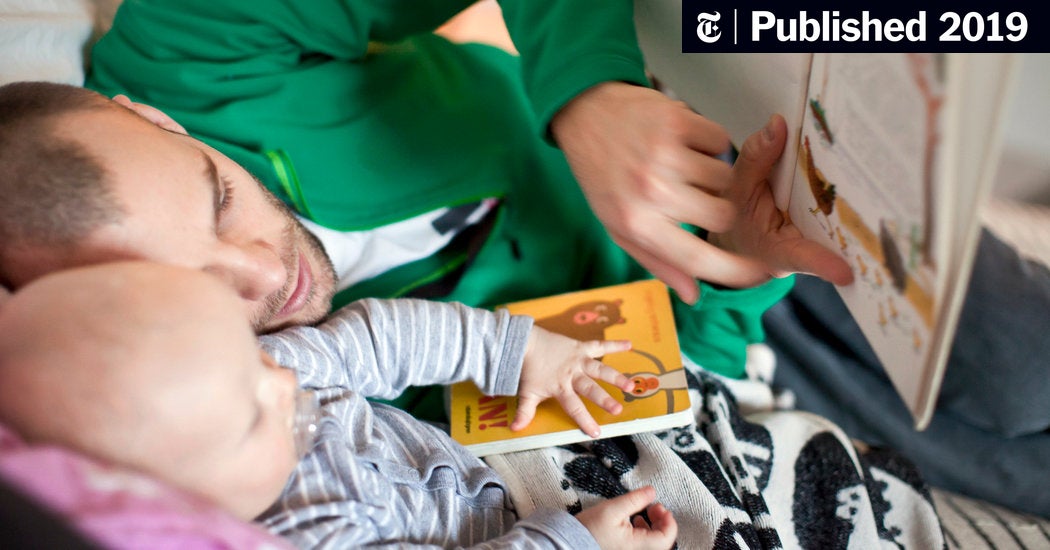}$}    & \multirow{3}{*}{\tabincell{l}{Reading to Your Toddler? \\ Print Books Are Better \\ Than Digital Ones}} & \multirow{3}{*}{Well}   \\
& & & & & & \\
& & & & & & \\\midrule
\multirow{3}{*}{False} & \multirow{3}{*}{True} & \multirow{3}{*}{False} & \multirow{3}{*}{2.40\%} & \multirow{3}{*}{$\includegraphics[width=0.12\textwidth]{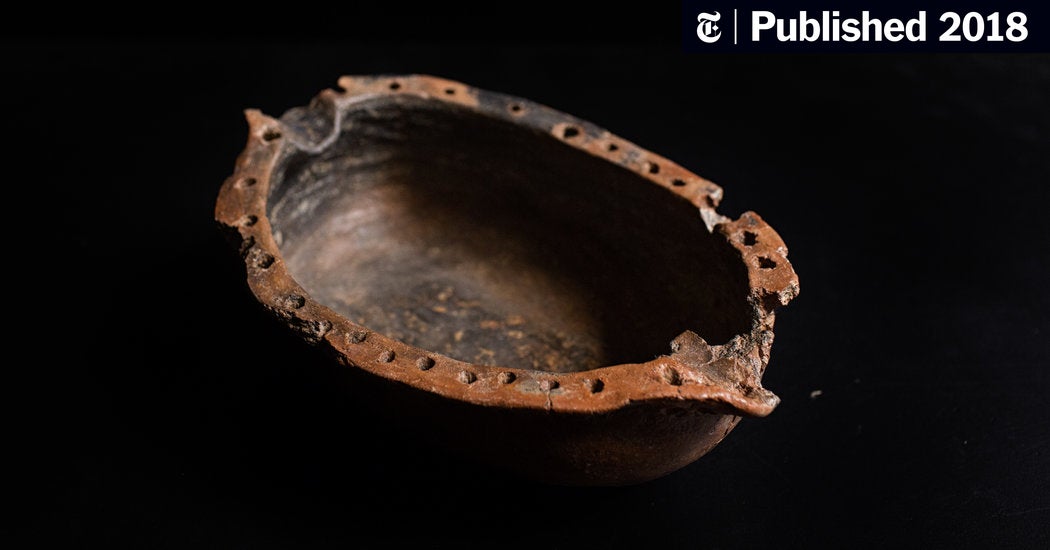}$}  & \multirow{3}{*}{\tabincell{l}{Italy's Oldest Instrument \\ Hints at Sounds of \\ Prehistoric Rome}}  & \multirow{3}{*}{Music}    \\
& & & & & & \\
& & & & & & \\\midrule
\multirow{3}{*}{False} & \multirow{3}{*}{False} & \multirow{3}{*}{False} & \multirow{3}{*}{14.22\%}  & \multirow{3}{*}{$\includegraphics[width=0.12\textwidth]{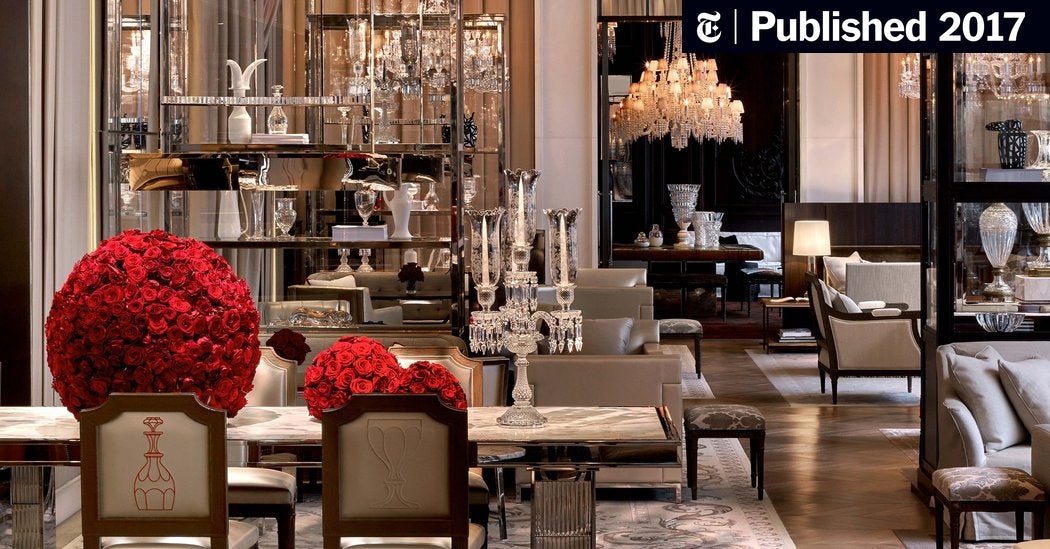}$}  & \multirow{3}{*}{\tabincell{l}{Tired of Mother's Day \\ Brunch? Try a Tea}} & \multirow{3}{*}{Travel}  \\
& & & & & & \\
& & & & & & \\\bottomrule
\end{tabular}
\caption{The experiment results with three types of network. \textit{True} means the classification is correct and \textit{False} means the classification is incorrect. For each type of prediction, an example of corresponding image-headline pair and its ground truth label is provided. And the \textit{Percent} represents each result ratio in testing set.}
\label{tab:ablation}
\end{table*}

\section{Experiments}


\subsection{Experimental Settings}

We trained all the models in the \textit{N24News} training set and the accuracy is tested on the testing set. Batch size is set to 32 and the learning rate is 1e-5 with an Adam \cite{kingma2014adam} optimizer. Each input image is resized to $224 \times 224$ and the maximum length of each input text is set to 512. Training device is an NVIDIA Tesla V100 with 16 GB RAM. For each training process, we train the model with the training set, retain the model that performs best on the validation set, and apply it on the testing set.

\subsection{Results}

All the experiment results are shown in Table \ref{tab:experiments}. We firstly classify the images and texts using ViT and RoBERTa respectively. In image classification task the accuracy is only 52.80 in F1, while RoBERTa behaves much better at the news text classification task. The lowest F1 is 70.31 using \textit{Headline} and the highest F1 is 87.65 with \textit{Body}.
There is a direct correlation between RoBERTa classification accuracy and text length.
From \textit{Headline}, \textit{Caption}, \textit{Abstract} to \textit{Body}, the longer the text length, the higher classification accuracy can be achieved. This is because RoBERTa can better understand the text with more meaningful words.
It is found that the multimodal classifier is better than either the image classifier or the text classifier. Even for the \textit{Body}, the improvement reaches 2.79 in F1 (\textbf{87.65} vs. \textbf{90.44}).
This is powerful proof that multimodal learning combining image features and text features benefits news classification. And the result also shows that the shorter the text (from body to headline), the more obvious the gain effect of adding image features. In other words, when text contains insufficient information, image is a perfect supplement.

\subsection{Error Analysis}

To explore why the multimodal method surpasses the text-only method, we separate the trained baseline model into three types: original multimodal network, image classification network with only the ViT, and text classification network with only the RoBERTa. We then test them in the testing dataset using image-headline pairs. The experimental results are shown in Table \ref{tab:ablation}. 

It is evident that when image and text are both correctly classified, the multimodal network can nearly always classify news correctly. The correct-to-incorrect ratio is \textbf{42.46}:\textbf{0.03}. Additionally, when image and text are both wrongly classified, the multimodal network also tends to be incorrect, but the correct-to-incorrect ratio is \textbf{14.22}:\textbf{2.56}, much lower than the previous \textit{Three True} situation. This shows that multimodal network can learn something useful after the features fusion of image and text, which may not be discovered if we process image and text separately.

Things are much more complex when only one of the image and the text classifiers is correct. The correct-to-incorrect ratio of multimodal classifier is (27.69+7.01=\textbf{34.7}):(2.40+3.63=\textbf{6.03}) in this situation. This shows that after proper training, the multimodal network will be more affected by the sub-network which can correctly perform the classification task. And this explains why our multimodal method is useful and able to outperform image-only and text-only networks.

The experiment results can be better understood by the examples in Table \ref{tab:ablation}. It can be observed that images and texts can provide some complementary information. The multimodal method can thus classify news more accurately. In the third row, the topic of news may be easily considered about \textit{Automobiles} if only considered the keyword \textit{Shift Gears} in text. But when considering the image, the scene described in this image obviously talks about the food, not the car. On the contrary, in the fourth row, a group of people are performing on the stage. It is hard to categorize whether this news article belongs to \textit{Dance} or \textit{Theater} without texts. Luckily, the headline directly tells us they are dancing, and this article must belong to \textit{Dance}.


Based on the above analysis, there are two main methods to further improve the performance of multimodal classification networks. The first one is to improve the behavior of each sub-network. If the accuracy (error) of sub-models is higher (lower), the multimodal prediction will also be improved. Our experimental results show that current state-of-the-art image classification models still have a long way to classify all news images correctly. The second method is to let the multimodal classifier be able to determine which sub-classifier extracts the more valuable feature. To do this, a more effective fusion network is needed to better combine image and text features.

\section{Conclusion}

In this paper, we introduce a multimodal news dataset \textit{N24News}, which is collected from the New York Times containing both images and texts, enables the multimodal research in real news classification. Compared to previous datasets, it covers almost all the essential news categories in our daily life, 
making the research on it more applicable to the real world. 
Based on \textit{N24News}, we propose a multitask multimodal network, which leverages the current state-of-the-art image classification model and text classification model. Experimental results show that combining image features and text features can achieve better classification accuracy comparing to the previous text-only methods. Our error analysis explains multimodal approach is helpful because the information of images and texts can complement each other. Accordingly, future work on improving the multimodal classification accuracy could include two main aspects: 1) improving image and text classification accuracy separately, especially the news image classification; 2) designing a more effective fusion network to better combine image and text features. 

\section{Bibliographical References}\label{reference}

\bibliographystyle{lrec2022-bib}
\bibliography{lrec2022-example}

\section{Language Resource References}
\label{lr:ref}
\bibliographystylelanguageresource{lrec2022-bib}
\bibliographylanguageresource{languageresource}

\end{document}